%% file: main.tex
\newif\ifreview
\reviewfalse       

\long\def\removedforreview#1{%
  \ifreview
    [removed for review]%
  \else
    #1%
  \fi
}

\documentclass[letterpaper]{article} 
\usepackage[preprint]{aaai2027}    
\usepackage[hyphens]{url}  
\usepackage{graphicx}      
\def\UrlFont{\rm}          
\usepackage{natbib}        
\usepackage{caption}       
\usepackage{algorithm}
\usepackage{algorithmic}
\usepackage{csquotes}
\usepackage{newfloat}
\usepackage{listings}
\DeclareCaptionStyle{ruled}{labelfont=normalfont,labelsep=colon,strut=off} 
\floatstyle{ruled}
\newfloat{listing}{tb}{lst}{}
\floatname{listing}{Listing}

\usepackage{booktabs}

\usepackage[utf8]{inputenc}
\usepackage[T1]{fontenc}
\usepackage{amsfonts}
\usepackage{amsmath}
\usepackage{amssymb}
\usepackage{microtype}
\usepackage{xcolor}
\usepackage{subcaption}
\usepackage{etoolbox}
\usepackage{siunitx}
\usepackage{booktabs}
\usepackage{multirow}
\usepackage{threeparttable}
\usepackage{rotating}
\usepackage{comment}

\providecommand{\href}[2]{#2 (\url{#1})}
\AtBeginEnvironment{figure}{\let\textwidth\columnwidth}

\title{MechaTerp-TRACE: A Novel Approach for Component Ablation Analysis in Language Models}

\author{
    Brandon Colelough\corresponding
    , Davis Bartels
    , Madeline Bittner
    , Dina Demner-Fushman
}
\affiliations{
    National Institutes of Health, National Library of Medicine\\
    Bethesda, Maryland, USA\\
}

\begin{document}
\maketitle

\begin{abstract}
\input{subpages/000_abstract}
\end{abstract}

\section{Introduction}
\input{subpages/001_introduciton}

\section{Background}
\input{subpages/002_background}

\section{Methods}
\input{subpages/003_methods}

\section{Results}
\input{subpages/004-a_results_figs_and_tables}
\input{subpages/004_results}

\section{Discussion}
\input{subpages/005_discussion}

\section{Conclusion}
\input{subpages/006_conclusion}

\section{Acknowledgments}
\input{subpages/011-Acknowledgment}

\bibliography{aaai2027}


\end{document}

%% file: subpages/000_abstract.tex
Interpretability research on large language models has produced accounts of factual recall in feed-forward layers and of token relationships in self-attention, but little work offers a unified way to compare the causal contribution of different architecture components to a model's output. We introduce MechaTerp (the \textbf{Mech}anistic In\textbf{terp}retability suite) -TRACE (subset for \textbf{T}eacher-forced \textbf{R}egistry of \textbf{A}blated \textbf{C}omponent \textbf{E}ffects), an architecture and study that measures how much each registered component of a language model supports the production of a named entity. TRACE ablates one component at a time and measures the resulting change in the output distribution at a fixed answer token, so component types from whole transformer blocks down to individual neurons and output logits can be compared on a common scale. We apply it to thirteen instruction-tuned dense decoder models spanning five families and one to thirty billion parameters, ablating 49,656 components across 48 medical and 42 general-knowledge prompts. We find that the components carrying the most effect are the same few, positionally fixed components in every model, regardless of which entity a prompt asks about, and that once these are removed the remaining support is close to evenly spread in eleven of the thirteen models. Apparent localisation of entity knowledge is therefore largely attributable to generic generation machinery, which has direct consequences for methods that assume entity knowledge sits in a findable place, including targeted knowledge editing.

%% file: subpages/001_introduciton.tex
Large Language Models (LLMs) are increasingly being deployed in high-stakes domains, where their outputs can directly influence consequential decisions. Despite their growing capabilities, these models remain largely “black-box” whose internal mechanisms are only partially understood, motivating a growing focus on mechanistic interpretability to understand how they operate \cite{zhao2024explainability, oberry2026quality}. As reliance on LLMs in these settings grows, low-quality or incorrect outputs carry increasingly detrimental consequences \cite{gallagher2024assessing}. However, these models remain prone to hallucinations, producing factually incorrect or fabricated answers, which highlights the need to look inside the model itself to understand why it arrives at a given answer \cite{huang2025hallucination}. Answering this requires knowing where knowledge is localized within the model. 


%% file: subpages/002_background.tex
Prior mechanistic interpretability research has largely relied on intervention methods such as activation patching to localize the internal components involved in specific model behaviors, specifically factual recall \cite{meng2022locating, geva2023dissecting, rai2024mechanistic}. These approaches identify activations at localized components whose restoration is sufficient to recover a target prediction, providing insight into where information relevant to factual recall is represented within a model. However, this insight is limited to the specific components tested in this method and does not generate a measure of a component's contribution that could be compared against different component types.  Separately, Templeton et al. used sparse autoencoders (SAEs) to learn feature representations from residual stream activations, demonstrating that it is possible to identify specific places in the architecture with correlative relationships to specific model behaviour \cite{Templeton2024Scaling}. This approach provides insight into how output behaviour may be attributed to specific blocks, but requires the training of another model, which is not scalable, and does not allow for analysis at different levels of granularity. Work on circuit discovery identifies the minimum graph of model components required for a specific computation. This method effectively attributes a specific process to a set of components \cite{3666122.3666841}. Circuit discovery is often done over computational processes that can be formally defined, such as addition. This allows for automatic evaluation of outputs, but little work has been done on less rigid domains. Automatic circuit discovery is also an intractable problem and is impractical for the size of contemporary models. Other related studies have used model editing to investigate how different transformer components contribute to factual knowledge editing. PMET (Precise Model Editing in a Transformer) optimizes hidden-state representations in both the MHSA (Multi-Head Self-Attention) and FFN (Feed Forward Network) sublayers when editing a fact, while restricting permanent weight updates to the FFN. Ablation experiments suggest that MHSA primarily encodes general knowledge extraction patterns while also storing a small amount of factual knowledge, while the FFN remains the primary component for modifying factual associations \cite{li2024pmet}. However, these methods are often applied to fixed layers and could benefit from a comprehensive understanding of the contribution of each layer or component in a specific domain.


%% file: subpages/003_methods.tex
\subsubsection{Research Questions}
\label{subsec:research-questions}
Through the conduct of the ablative study, MechaTerp-TRACE, realised through the MechaTerp architecture, we aim to determine \textbf{to what extent is knowledge in a Language Model about a defined named entity localized to a discrete part of a model's internal architecture?} 

\subsubsection{Contributions}
\label{subsec:contributions}
\begin{itemize}
    \item A scalable framework (MechaTerp-TRACE) that determines entity-knowledge localization across any dense model family. 
    \item A behavioural experimentation apparatus, including a curated dataset of paired medical and general-knowledge question-answering prompts, used to elicit and observe model behaviour for the study.
    \item A large-scale ablative experiment and its results, providing the first broad evidence for whether entity knowledge is localized or spread throughout a model.
\end{itemize}

\subsection{MechaTerp-TRACE Design}
\label{sec:mechaterp-trace-design}

\subsection{Behaviouralistic Design}
\label{subsec:behaviouralistic-design}
Three major sub-components are present within the behavioural side of the MechaTerp-TRACE study, including the desired generation effect, the domain of the prompts used for named-entity elicitation, and the language model prompting methodology. We design the behavioural study around prompting the model to generate a single string of tokens that correspond directly to a named entity. For this reason, we focus only on dense language models and do not extend this study to \enquote{reasoning} or \enquote{thinking} models, as their \enquote{reasoning} trace may correspond to components within the language model that are not that of the named entity. To achieve our goal of eliciting a single named entity from a model, we develop a dataset of Question-Answer (QA) prompts wherein the answer to the provided prompt is a singular named entity. We center the prompts designed for the MechaTerp-TRACE study on two disparate domains to further measure the effect of commonly required generation components and domain-specific generation components. We define commonly required generation components to be \textbf{components of the model that support answer production across every prompt domain under study}, and domain-specific generation components to be \textbf{components of the model that support answer production within one prompt domain and not the others}. We select the medical and general knowledge domains to further study component importance across a specialised body of knowledge and a broad body of knowledge. Within the medical domain, we narrow to the drug subdomain and select eight drug entities including  Metformin, Acetaminophen, Semaglutide, Amoxicillin, Levodopa, Methylphenidate, Rosuvastatin and Furosemide, and for each entity we write one prompt under each of six clue categories including History, Chemical Formula, Generic Name, First-Line Treatment, Drug-Drug Interaction and Other.  for a total of 48 medical prompts. For the general knowledge domain, we draw on the TREC-8 \footnote{https://pages.nist.gov/trec-browser/trec8/qa/data/} question answering set, which
contains 200 questions. We reword each question so that its answer is a single named entity, matching the elicitation format used for the medical prompts. We then retain only those prompts for which the models under study produce the correct named entity,
working to a target of at least 30 retained prompts per domain so that both domains support comparison at a similar scale. This left a final general corpus of 42 prompts. We retained a question only when a majority of the models tested produced the correct named entity, so that any single model under study would answer at least 30 of the retained set correctly. Only 42 of the 200 TREC-8 questions met that condition. The last design detail for the behavioral side of the MechaTerp-TRACE architecture is the prompting methodology used. For the full ablative MechaTerp-TRACE study, we adopt a \textbf{continuation-style} prompting regime. We define continuation-style as \textbf{the placement of the prompt inside the model's own assistant turn, so that the model treats the prompt as text it has already begun producing and continues it, rather than as a question it has been asked to answer}. Across all model families tested, we render the chat template with the task instruction in the system turn, leave the user turn empty, open the assistant turn, and place the continuation stem immediately after the assistant header. The sequence is not closed with an end-of-turn marker, so from the model's perspective its own reply is already underway, and the natural next action is to continue the sentence. We adopt this prompting strategy so that the named entity always appears at a known position. 

\subsection{Mechanistic Design}
\label{sec:mechanistic-design}

We work on one model and one lane at a time. We define a lane as a group of components within a single model that can sensibly be compared with each other, and every calculation in this section happens inside a single lane. Let $\mathcal{U}$ be the set of components measured in a lane and $B$ denote the number of components per lane. Prompts are indexed as $r$, and components by $s$ and $j$. Normalisation is lane-internal, so each lane carries its own denominator. 

\subsubsection{Component Definition}
We define a component as a site within the model that can be removed from the forward pass on its own, leaving the rest of the computational process intact and still able to produce a logit vector at the answer anchor. Subsequently, the measured degradation of the answer-anchor logit vector under single-component ablation is the quantity we use to determine component importance for the generation of a specified named entity. We subcategorize model components into three families, including coarse structural sites, extended sites, and architecture-specific sites. We define each of these families as follows. \textbf{Coarse structural sites} are the components that make up the residual skeleton of the network and are therefore present in every dense decoder we test. These include the transformer block itself, the attention and MLP branches that write into the residual stream, the residual carrier those branches write to, and the individual attention heads. \textbf{Extended sites} are the parameters and dimensions that sit inside coarse site components and determine what that site computes. Extended sites include the query, key, value, and output projections of attention, the gate, up and down projections of the MLP, normalisation scale and shift parameters, layer biases, positional patches, and rows, channels, and logits of the output head. \textbf{Architecture-specific sites} are components that exist only where a model family provides the mechanism, such as expert routing in mixture-of-experts
layers, sliding-window and hybrid attention surfaces, fused query-key-value projections, and recurrent or state-space memory.

\subsubsection{Component Ablation}
Component ablation is dependent upon component type and unique to each component ablated. MechaTerp-TRACE supports considerably more ablation operations than can be listed here, and the complete operator table is published with the implementation source code at \removedforreview{\url{https://github.com/Brandonio-c/ClinIQLink-MechaTerp}}. The goal at every level of component ablation is to remove the contribution of the target component as completely as we can while reducing residual effects to surrounding components, so that the change measured at the answer anchor is attributable to that component and not to collateral disruption of the surrounding computation. At the coarsest level of ablation, a transformer block can be bypassed outright, since the block returns a hidden state of the same shape it received, so substituting the input for the output removes the block's entire net update. Components that write into a residual carrier cannot be handled the same way, and as such, the ablation method for components such as attention and MLP branches contributes additively, so the correct removal is to zero the branch contribution before it is written back, leaving the carrier itself intact. Every component ablation measurement is taken at a single token position, which we call the answer anchor. The anchor is the position at which the model is about to produce the first token that carries the named entity. We fix the anchor by teacher forcing, meaning we supply the correct answer tokens as context rather than letting the model generate them. We render the continuation-style prompt, append the correct answer text, resolve the first informative answer token, and truncate the tokenisation immediately before it, so a single forward pass yields the logits at the position where the entity would be predicted. Continuation-style prompting constrains the entity to the next generated position and teacher forcing pins that position to a fixed index taken from the prompt package. 

\subsubsection{The Ablation-Effect Metric}

Let $z^{(0)}_{r}$ be the intact logit vector at the answer anchor and $z^{(-s)}_{r}$ the same vector after ablating site $s$. Our primary effect metric is the cosine distance between them: 
\begin{equation}
\Delta_{r,s} = 1 - \cos\!\left(z^{(0)}_{r},\, z^{(-s)}_{r}\right)
\;\in\; [0,2].
\label{eq:delta}
\end{equation}
Every ablated vector is compared back to the same intact vector, which keeps all components on a common reference scale. Equation~\eqref{eq:delta} measures rotation of the logit vector, and the softmax is invariant to $z \mapsto z + c\mathbf{1}$, which \eqref{eq:delta} is not, and \eqref{eq:delta} is invariant to $z \mapsto \alpha z$ for $\alpha>0$, which the softmax is not. 

\subsubsection{Effect Weights and Lane Shares}

Since $\Delta_{r,s} \ge 0$, each component contributes a nonnegative weight $w_{r,s} = \max(0, \Delta_{r,s})$. Normalising by the lane total gives shares that sum to one, so profiles can be compared across prompts of differing total effect,
\begin{equation}
W_{r,\ell} = \sum_{j \in \mathcal{U}_{m,\ell}} w_{r,j},
\qquad
p_{r,s\mid\ell} = \frac{w_{r,s}}{W_{r,\ell}},
\qquad W_{r,\ell} > 0 .
\label{eq:shares}
\end{equation}

where $w_{r,s}$ is the effect weight of component $s$ under prompt $r$, $W_{r,\ell}$ is the total effect measured across lane $\ell$, and $p_{r,s\mid\ell}$ is the share of that total carried by component $s$.

\subsection{Knowledge Localization}
For a single prompt, ablating each component in turn yields one nonnegative effect per component within a fixed lane universe. We call this vector of effects the prompt's profile, and we summarise each profile by measuring how much of the total effect is concentrated over the ablated components. A small set of components carrying most of the effect indicates concentrated support, while an even spread indicates distributed support.

\subsubsection{Concentration, Localisation, and Effective Support}
To measure the component importance concentration profile, we adopt a modified version of the Herfindahl-Hirschman Index~\cite{hirschman1964paternity}, computed over lane shares. For $W_{r,\ell} > 0$,

\begin{equation}
\operatorname{HHI}_{r,\ell} = \sum_{s \in \mathcal{U}_{m,\ell}} p_{r,s\mid\ell}^{\,2}
\;\in\; \left[\tfrac{1}{B_{m,\ell}},\, 1\right],
\label{eq:hhi}
\end{equation}
and for $B_{m,\ell} > 1$,
\begin{equation}
\operatorname{LLI}_{r,\ell} =
\frac{B_{m,\ell}\operatorname{HHI}_{r,\ell} - 1}{B_{m,\ell} - 1},
\qquad
N_{\mathrm{eff},\,r,\ell} = \frac{1}{\operatorname{HHI}_{r,\ell}}.
\label{eq:lli}
\end{equation}
where $\operatorname{HHI}_{r,\ell}$ is the concentration of effect for prompt $r$ in lane $\ell$, $p_{r,s\mid\ell}$ is the share of that lane's effect carried by component $s$, and $B_{m,\ell}$ is the number of components in the lane universe $\mathcal{U}_{m,\ell}$. Squaring the shares rewards large shares disproportionately, so the score is $1/B_{m,\ell}$ when all components contribute equally and reaches $1$ when a single component carries all the effect of the generation process. $\operatorname{LLI}_{r,\ell}$ (Lane Localisation Index) rescales $\operatorname{HHI}_{r,\ell}$ to run between zero and one regardless of lane size, and $N_{\mathrm{eff},\,r,\ell}$ (number of effective components) inverts $\operatorname{HHI}_{r,\ell}$ to give the number of equally contributing components that would produce the observed concentration.

\subsubsection{Cross-Prompt Profile Overlap}

$\operatorname{LLI}_{r,\ell}$ measures the concentration of component importance within a single prompt. To measure the recurrence of component importance across prompts targeting a specified named entity, we compare profiles pairwise using a weighted Jaccard overlap. For two prompts sharing an aligned lane universe, both with positive effect mass, we compute: 
\begin{equation}
J_{\ell}(r,r') =
\frac{\sum_{s} \min\!\left(p_{r,s\mid\ell},\, p_{r',s\mid\ell}\right)}
     {\sum_{s} \max\!\left(p_{r,s\mid\ell},\, p_{r',s\mid\ell}\right)}.
\label{eq:jaccard}
\end{equation}

where $r$ and $r'$ are two prompts measured over the same lane $\ell$, and $p_{r,s\mid\ell}$ and $p_{r',s\mid\ell}$ are the shares of lane effect that component $s$ carries in each.  $J_{\ell}$ is measured from zero when no component is used by both prompts to one when the two profiles are identical.

\subsection{Generation Floor}

We define the generation floor as the set of components whose ablation effect, measured across a broad and content-diverse sample of prompts, is (a) consistently high in magnitude and (b) consistently low in variance. Generation floor components always share a high concentration of component importance to the generation process regardless of the prompt used by the model. We measure the generation floor utilising the general knowledge QA dataset, as the named entity for each prompt is different for each prompt and, as such, a component that stays important
throughout is unlikely to be important because of any particular entity.

\subsubsection{Domain-Common Components}

Let $R_{M}$ and $R_{G}$ be the medical and general prompt sets, with absent
prompt-site pairs contributing zero,
\begin{equation}
\bar{p}^{\,M}_{s\mid\ell} =
\frac{1}{|R_{M}|}\sum_{r \in R_{M}} p_{r,s\mid\ell},
\label{eq:domain-mean}
\end{equation}
and $\bar{p}^{\,G}_{s\mid\ell}$ defined analogously over $R_{G}$. Let $q^{M}_{s\mid\ell}$ and $q^{G}_{s\mid\ell}$ be the fraction of prompts in each family with $p_{r,s\mid\ell} \ge \tau_{p}$. Commonness is harmonic, so it is high only when both domain means are high,

\begin{equation}
h_{s\mid\ell} =
\frac{2\,\bar{p}^{\,M}_{s\mid\ell}\,\bar{p}^{\,G}_{s\mid\ell}}
     {\bar{p}^{\,M}_{s\mid\ell} + \bar{p}^{\,G}_{s\mid\ell} + \varepsilon},
\qquad
\tilde{h}_{s\mid\ell} =
\frac{h_{s\mid\ell}}{\max_{j \in \mathcal{U}_{m,\ell}} h_{j\mid\ell} + \varepsilon},
\label{eq:commonness}
\end{equation}

A component site is found to be domain-common when:

\begin{equation}
\begin{split}
C^{\mathrm{com}}_{s\mid\ell} = \mathbf{1}\Big\{\,
  &q^{M}_{s\mid\ell} \ge q_{\min} \;\wedge\;
   q^{G}_{s\mid\ell} \ge q_{\min} \\
  &\wedge\; \bar{p}^{\,M}_{s\mid\ell} \ge \tau_{\mathrm{share},\ell}
   \;\wedge\; \bar{p}^{\,G}_{s\mid\ell} \ge \tau_{\mathrm{share},\ell}
\,\Big\},
\end{split}
\label{eq:common-flag}
\end{equation}
with $\tau_{\mathrm{share},\ell}$ the lane's 75th percentile of the mean of the two domain means.

\subsubsection{The Generation Floor Correction}

Let $\mu_{s\mid\ell}$ and $\operatorname{occ}_{s\mid\ell}$ be the mean share and occurrence frequency of site $s$ over $R_{G}$, computed as in \eqref{eq:domain-mean}, with quartiles taken across the same prompts. Stability is the coefficient of quartile variation,

\begin{equation}
\operatorname{CQV}_{s\mid\ell} =
\frac{Q_{3,s\mid\ell} - Q_{1,s\mid\ell}}
     {Q_{3,s\mid\ell} + Q_{1,s\mid\ell} + \varepsilon}.
\label{eq:cqv}
\end{equation}
A component site is found to be a generation floor candidate when

\begin{equation}
\begin{split}
C^{\mathrm{flr}}_{s\mid\ell} = \mathbf{1}\Big\{\,
  &\operatorname{occ}_{s\mid\ell} \ge \omega_{\min} \;\wedge\;
   \mu_{s\mid\ell} \ge \tau_{\mu,\ell} \\
  &\wedge\; \operatorname{CQV}_{s\mid\ell} \le \tau_{\mathrm{stab},\ell}
\,\Big\},
\end{split}
\label{eq:floor-flag}
\end{equation}

with $\tau_{\mu,\ell}$ the lane's 75th percentile of mean shares and $\tau_{\mathrm{stab},\ell}$ its 25th percentile of $\operatorname{CQV}$. Candidates are scored by combining magnitude with stability, so that a component scores highly only when the effect is both large and consistent across the general prompts,

\begin{equation}
f_{s\mid\ell} = \mu_{s\mid\ell}\left(1 - \operatorname{CQV}_{s\mid\ell}\right),
\qquad
\tilde{f}_{s\mid\ell} =
\frac{f_{s\mid\ell}}
     {\max_{j:\,C^{\mathrm{flr}}_{j\mid\ell}=1} f_{j\mid\ell} + \varepsilon},
\label{eq:floor-score}
\end{equation}

where $f_{s\mid\ell}$ is the raw floor score for component $s$ and $\tilde{f}_{s\mid\ell}$ rescales it against the strongest flagged candidate in the lane, so scores from lanes of differing size remain comparable.

%% file: subpages/004-a_results_figs_and_tables.tex
\input{subpages/results_combined_no_gemma}

\begin{figure}[t]
\centering
\includegraphics[width=\linewidth]{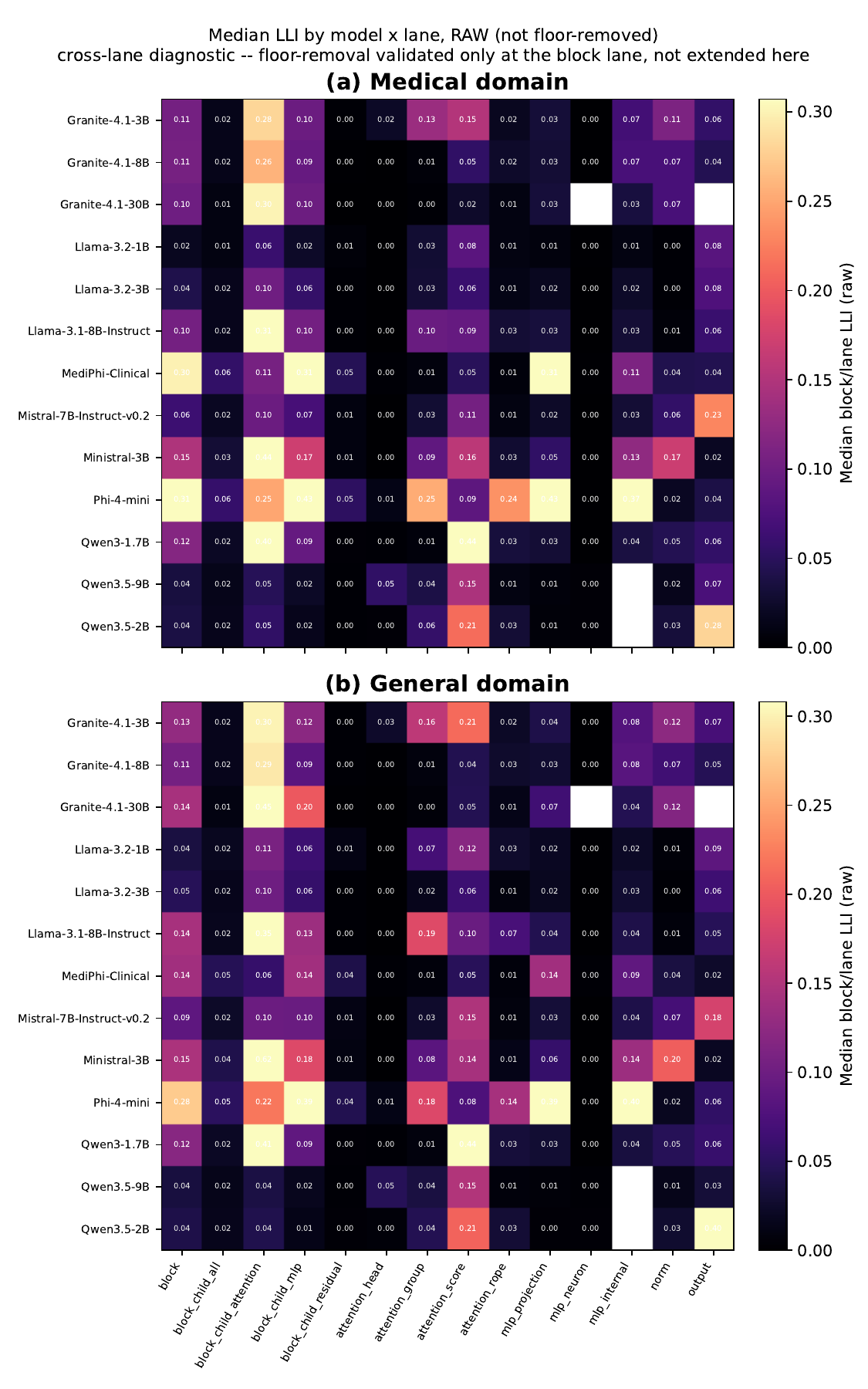}
\caption{Median block/lane LLI (Eq.~\ref{eq:lli}) by model and component lane, \emph{before} floor removal (raw), split by domain. Shown as a cross-lane diagnostic: floor removal is validated and applied only at the block lane (Fig.~\ref{fig:neff-ranked} and Table~\ref{tab:results-combined-no-gemma}), so the raw values here illustrate where effect concentrates architecturally prior to correction and should not be read as the corrected localization result. Blank cells denote lanes absent from that model's architecture. Sub-figure A shows results for medical prompt subset and sub-figure B shows results for general-knowledge prompt subset.}
\label{fig:lane-heatmap}
\end{figure}

\begin{figure}[t]
\centering
\includegraphics[width=\linewidth]{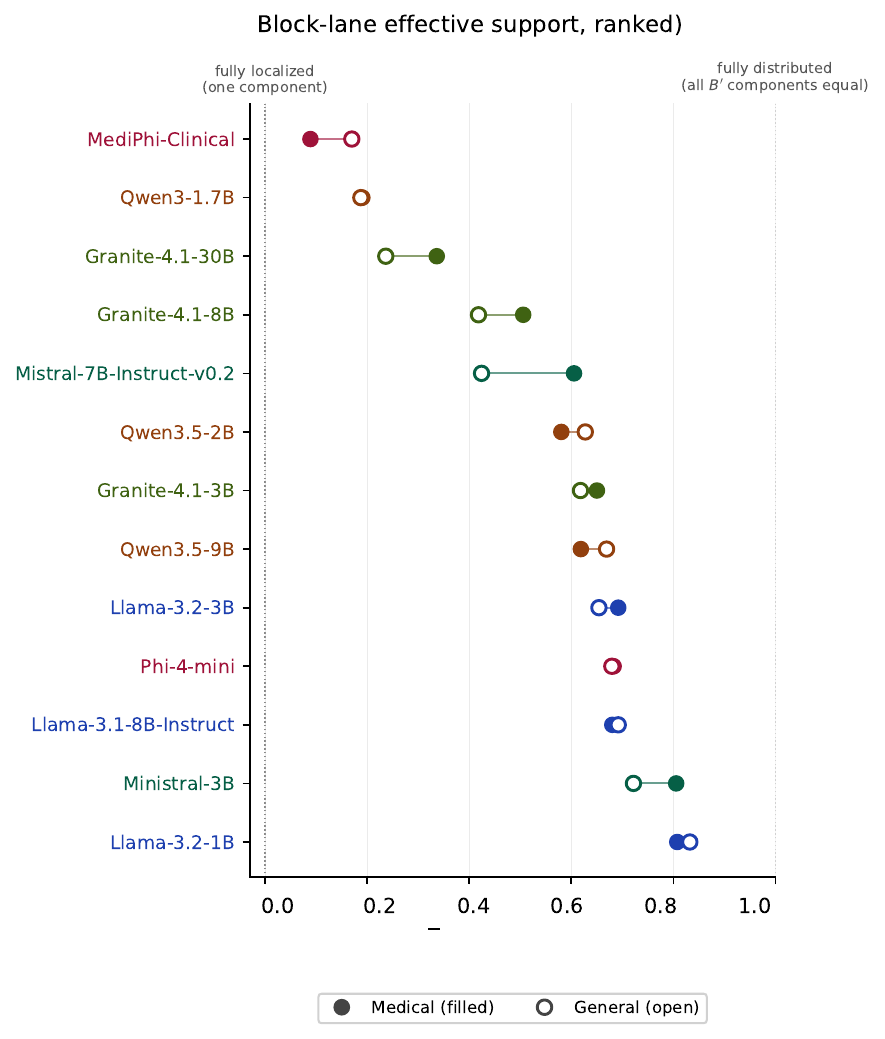}
\caption{Block-lane effective support ($N_{\mathrm{eff}}/B'$, Eq.~\ref{eq:lli}) after hard removal of generation-floor components (Eq.~\ref{eq:floor-flag}), ranked across the 13-model cohort. Lower values indicate more localized entity support (few components carry the effect); higher values indicate distributed support. Once structurally-fixed always-on components are removed, all models but MediPhi-Clinical and Qwen3-1.7B fall toward the distributed pole, indicating that raw concentration is largely attributable to generic generation machinery rather than entity-specific processing. Filled markers: medical prompts; open markers: general prompts.}
\label{fig:neff-ranked}
\end{figure}

%% file: subpages/results_combined_no_gemma.tex
\begin{table*}[t]
\centering
\captionsetup{font=small}
\caption{MechaTerp-TRACE per-model results (13 models; Gemma excluded from this cohort), combining localisation (Eq. 4/5, hard-floor-removed -- see notes), cross-prompt reuse (Eq. 6), and domain-common/generation-floor components with their soft LLI correction (Eq. 9, 11, 12), transposed with models as columns. Short labels (family-Params): Gr-3B=Granite-4.1-3B, Gr-8B=Granite-4.1-8B, Gr-30B=Granite-4.1-30B, L-1B=Llama-3.2-1B, L-3B=Llama-3.2-3B, L-8B=Llama-3.1-8B-Instruct, Mi-7B=Mistral-7B-Instruct-v0.2, Mn-3B=Ministral-3B, MP=MediPhi-Clinical, Ph=Phi-4-mini, Q-1.7B=Qwen3-1.7B, Q3.5-9B=Qwen3.5-9B, Q3.5-2B=Qwen3.5-2B. }
\label{tab:results-combined-no-gemma}
\footnotesize
\setlength{\tabcolsep}{2.5pt}
\begin{threeparttable}
\begin{tabular}{lrrrrrrrrrrrrr}
\toprule
{} & \multicolumn{3}{c}{Granite} & \multicolumn{3}{c}{Llama} & \multicolumn{2}{c}{Mistral} & \multicolumn{2}{c}{Phi} & \multicolumn{3}{c}{Qwen} \\
\cmidrule(lr){2-4}\cmidrule(lr){5-7}\cmidrule(lr){8-9}\cmidrule(lr){10-11}\cmidrule(lr){12-14}
{Metric} & Gr-3B & Gr-8B & Gr-30B & L-1B & L-3B & L-8B & Mi-7B & Mn-3B$^\dagger$ & MP & Ph & Q-1.7B & Q3.5-9B & Q3.5-2B \\
\midrule
\multicolumn{14}{l}{\textbf{Identity}} \\
\midrule
Params & 3B & 8B & 30B & 1B & 3B & 8B & 7B & 3B & 3.8B & 3.8B & 1.7B & 9B & 2B \\
$B$ (block-lane size) & 40 & 40 & 64 & 16 & 28 & 32 & 32 & 14 & 32 & 32 & 28 & 32 & 24 \\
\midrule
\multicolumn{14}{l}{\textbf{Correctness}} \\
\midrule
Correct (of 90) & 63 & 75 & 77 & 46 & 71 & 68 & 72 & 0 & 71 & 64 & 35 & 57 & 55 \\
Med. correct (of 48) & 29 & 36 & 38 & 14 & 35 & 28 & 33 & 0 & 37 & 28 & 12 & 24 & 31 \\
Gen. correct (of 42) & 34 & 39 & 39 & 32 & 36 & 40 & 39 & 0 & 34 & 36 & 23 & 33 & 24 \\
\midrule
\multicolumn{14}{l}{\textbf{Localisation (block lane, floor-removed)}} \\
\midrule
$B'$ (floor-removed) & 33 & 33 & 57 & 13 & 26 & 27 & 28 & 12 & 30 & 26 & 26 & 30 & 23 \\
LLI blk med (median) & 0.017 & 0.030 & 0.035 & 0.020 & 0.018 & 0.018 & 0.024 & 0.022 & 0.354 & 0.019 & 0.170 & 0.021 & 0.033 \\
LLI blk gen (median) & 0.019 & 0.043 & 0.058 & 0.017 & 0.021 & 0.017 & 0.050 & 0.035 & 0.169 & 0.019 & 0.173 & 0.017 & 0.027 \\
$N_{eff}$ med & 21.5 & 16.7 & 19.2 & 10.5 & 18.0 & 18.4 & 16.9 & 9.7 & 2.7 & 17.8 & 5.0 & 18.6 & 13.3 \\
$N_{eff}$ gen & 20.4 & 13.8 & 13.5 & 10.8 & 17.0 & 18.7 & 11.9 & 8.7 & 5.1 & 17.7 & 4.9 & 20.1 & 14.4 \\
$N_{eff}/B'$ & 0.63 & 0.46 & 0.29 & 0.82 & 0.67 & 0.69 & 0.51 & 0.76 & 0.13 & 0.68 & 0.19 & 0.64 & 0.60 \\
LLI correct & 0.019 & 0.038 & 0.043 & 0.015 & 0.019 & 0.019 & 0.036 & {--} & 0.207 & 0.018 & 0.176 & 0.022 & 0.031 \\
LLI incorrect & 0.015 & 0.030 & 0.036 & 0.021 & 0.019 & 0.017 & 0.021 & 0.026 & 0.393 & 0.019 & 0.165 & 0.017 & 0.029 \\
\midrule
\multicolumn{14}{l}{\textbf{Cross-prompt reuse (block lane)}} \\
\midrule
Jac within (median) & 0.785 & 0.803 & 0.749 & 0.796 & 0.777 & 0.758 & 0.761 & 0.906 & 0.512 & 0.756 & 0.868 & 0.785 & 0.762 \\
Jac across (median) & 0.730 & 0.741 & 0.670 & 0.723 & 0.696 & 0.732 & 0.704 & 0.865 & 0.385 & 0.710 & 0.875 & 0.721 & 0.698 \\
$\Delta$Jac & 0.055 & 0.062 & 0.079 & 0.072 & 0.081 & 0.026 & 0.056 & 0.040 & 0.127 & 0.046 & $-$0.007$^{\mathrm{R}}$ & 0.064 & 0.065 \\
$p$ (Mann-Whitney $U$) & $<$.001 & $<$.001 & $<$.001 & $<$.001 & $<$.001 & $<$.001 & $<$.001 & $<$.001 & $<$.001 & $<$.001 & 0.983$^{\mathrm{R}}$ & $<$.001 & $<$.001 \\
\midrule
\multicolumn{14}{l}{\textbf{Generation floor (block lane)}} \\
\midrule
$N_{com}$ & 10 & 10 & 15 & 3 & 5 & 6 & 7 & 3 & 7 & 5 & 7 & 6 & 4 \\
$N_{flr}$ & 7 & 7 & 7 & 3 & 2 & 5 & 4 & 2 & 2 & 6 & 2 & 2 & 1 \\
$\Delta$LLI & $-$0.044 & $-$0.044 & $-$0.052 & $-$0.009 & $-$0.020 & $-$0.050 & $-$0.031 & $-$0.071 & 0.013 & $-$0.045 & 0.012 & $-$0.016 & $-$0.012 \\
$K_{50}$ & 3.0 & 3.0 & 3.0 & 5.0 & 5.0 & 3.0 & 4.0 & 2.0 & 2.0 & 2.0 & 3.0 & 7.0 & 5.0 \\
$K_{80}$ & 9.0 & 9.0 & 14.0 & 10.0 & 15.0 & 10.0 & 12.5 & 6.0 & 8.0 & 3.0 & 10.0 & 17.0 & 13.0 \\
$N_{com \cap flr}$ & 7 & 6 & 7 & 3 & 2 & 5 & 4 & 2 & 2 & 5 & 2 & 2 & 1 \\
Overlap frac. & 0.700 & 0.545 & 0.467 & 1.000 & 0.400 & 0.833 & 0.571 & 0.667 & 0.286 & 0.833 & 0.286 & 0.333 & 0.250 \\
\midrule
\multicolumn{14}{l}{\textbf{Format compliance (general domain, of 42)}} \\
\midrule
Compliant rate & 1.000 & 1.000 & 1.000 & 1.000 & 1.000 & 1.000 & 0.881 & 0.190 & 1.000 & 1.000 & 1.000 & 0.810 & 0.976 \\
No answer & 0 & 0 & 0 & 0 & 0 & 0 & 0 & 0 & 0 & 0 & 0 & 8 & 0 \\
Verbose & 0 & 0 & 0 & 0 & 0 & 0 & 5 & 34 & 0 & 0 & 0 & 0 & 1 \\
\bottomrule
\end{tabular}
\begin{tablenotes}[flushleft]
\scriptsize
\setlength{\itemsep}{0pt}\setlength{\parskip}{0pt}\setlength{\topsep}{0pt}
\item Metrics per Eqs.~5, 6, 9, 11, 12; medians over prompts (block lane). $N_{eff}/B$ and $K_{50}/K_{80}$ pool domains.
\item $\dagger$ Mn-3B: correctness-conditioned rows reflect format non-engagement, not knowledge (it rephrases/repeats rather than answering). All other rows are correctness-independent. \{--\}: no correct prompts, so LLI$_{\text{correct}}$ undefined.
\item $^{\mathrm{R}}$ Q-1.7B: $\Delta$Jac reversed and non-significant ($p\geq.05$); all others $\Delta$Jac$>0$, $p<.001$ ($p<.001$ shown as $<$.001).
\item $\Delta$LLI $<0$: removing always-on components lowers apparent localisation. Format compliance (general domain): compliant $=$ non-empty, $\leq$6 words, not repetitive; no answer $=$ empty; verbose $=$ $>$6 words (repetition category omitted, zero for all).
\item Head-lane LLI omitted: near-zero ($\sim$0.0004--0.06) across the cohort.
\end{tablenotes}
\end{threeparttable}
\end{table*}

%% file: subpages/004_results.tex
We ran sixteen models through the MechaTerp-TRACE architecture, though we report results for thirteen as the Gemma-3 family of models was not able to effectively produce results. Across the thirteen reported models the registry materialised 49,656 components in total, ranging from 950 for Qwen3.5-2B to 7,362 for Granite-4.1-30B, and every one of these was ablated individually against all 90 prompts. These components span fourteen lanes covering the whole architecture, namely the transformer block, its attention, MLP and residual children and their combination, individual attention heads, query groups, attention scores and positional encodings, MLP projections, internals and individual neurons, normalisation parameters, and the output head with its rows, channels and logits. The cohort was chosen to give reasonable coverage of current open-weight decoder-only models, spanning five families and a range of parameter counts from 1B to 30B, so that the MechaTerp-TRACE method could be shown to work across architectures and across scales. Every model tested is instruction tuned, which is preferable for the continuation-style prompting regime, since the prompt is placed inside an opened assistant turn. From the Granite family we tested Granite-4.1-3B, Granite-4.1-8B and Granite-4.1-30B. From the Llama family we tested Llama-3.2-1B, Llama-3.2-3B and Llama-3.1-8B-Instruct. From Mistral we tested Mistral-7B-Instruct-v0.2 and Ministral-3B. From the Phi family we tested Phi-4-mini and a fine-tuned model, MediPhi-Clinical, included because it is domain specialised for clinical text and offers a useful contrast against general-purpose models on the medical prompts. From the Qwen family we tested Qwen3-1.7B, Qwen3.5-2B and Qwen3.5-9B. The remaining three models were Gemma-3 at 1B, 4B and 12B, accounting for a further 7,836 components. Gemma-3 did not follow the short-answer instruction at any size, and it could not reliably produce the correct named entity for the prompt packages in either domain. Because MechaTerp-TRACE measures the effect of ablation at the position where the answer token should appear, a model that never reaches that answer under any condition yields profiles we cannot interpret.

Figure~\ref{fig:lane-heatmap} reports median LLI for each model and lane before generation floor removal. Values span 0.00 to 0.62 across the cohort. The highest single value is Ministral-3B at the \texttt{block\_child\_attention} lane on general prompts (0.62). That lane also holds the largest value for Granite-4.1-3B, Granite-4.1-8B, Granite-4.1-30B, Llama-3.1-8B-Instruct, Ministral-3B and Qwen3-1.7B in at least one domain, reaching 0.45 for Granite-4.1-30B and 0.41 for Qwen3-1.7B on general prompts. MediPhi-Clinical and Phi-4-mini instead take their largest values at the \texttt{block}, \texttt{block\_child\_mlp} and \texttt{mlp\_projection} lanes, at 0.30, 0.31 and 0.31 for MediPhi-Clinical on medical prompts and 0.31, 0.43 and 0.43 for Phi-4-mini. 

Figure~\ref{fig:neff-ranked} reports block-lane effective support ($N_{\mathrm{eff}}/B'$) after hard removal of generation-floor components, ranked from lowest to highest. Values span roughly 0.07 to 0.85. MediPhi-Clinical is lowest at approximately 0.07 on medical prompts and 0.16 on general prompts, followed by Qwen3-1.7B at approximately 0.17. Granite-4.1-30B, Granite-4.1-8B and Mistral-7B-Instruct-v0.2 occupy the middle of the range, at approximately 0.22 to 0.33, 0.41 to 0.50 and 0.42 to 0.61 respectively. The remaining eight models fall between approximately 0.58 and 0.85. Table~\ref{tab:results-combined-no-gemma} reports per-model results across the thirteen-model cohort. Block-lane size $B$ (the number of transformer blocks) ranges from 14 for Ministral-3B to 64 for Granite-4.1-30B. Correct answers out of 90 prompts range from 35 for Qwen3-1.7B  to 77 for Granite-4.1-30B, , with Ministral-3B returning none. The Median floor-removed block-lane LLI sits between 0.017 and 0.035 on medical prompts for eleven of the thirteen models. The two exceptions are MediPhi-Clinical, at 0.354 medical and 0.169 general, and Qwen3-1.7B, at 0.170 and 0.173. Effective support $N_{\mathrm{eff}}$ spans 2.7 to 21.5 on medical prompts and 4.9 to 20.1 on general prompts, and the pooled ratio $N_{\mathrm{eff}}/B'$ spans 0.13 for MediPhi-Clinical to 0.82 for Llama-3.2-1B. Splitting by baseline correctness, LLI is higher on incorrect prompts than on correct prompts for MediPhi-Clinical (0.393 against 0.207) and Llama-3.2-1B (0.021 against 0.015), and lower or equal for the remaining models. Median within-entity weighted Jaccard ranges from 0.512 for MediPhi-Clinical to 0.906 for Ministral-3B, and median across-entity overlap from 0.385 to 0.875. Between three and fifteen components are flagged domain-common per model and between one and seven as generation-floor candidates, with the intersection accounting for between 0.250 and 1.000 of the domain-common set. Removing floor components lowers median LLI for eleven models, by between 0.009 and 0.071, and raises it for MediPhi-Clinical (0.013) and Qwen3-1.7B (0.012). $K_{50}$ ranges from 2.0 to 7.0 components and $K_{80}$ from 3.0 to 17.0. Format compliance on the general domain is 1.000 for nine models, 0.976 for Qwen3.5-2B, 0.881 for Mistral-7B-Instruct-v0.2, 0.810 for Qwen3.5-9B, and 0.190 for Ministral-3B, whose 34 non-compliant responses were all verbose. Qwen3.5-9B accounts for the only empty responses in the cohort, at eight.

%% file: subpages/005_discussion.tex
\subsection{Composition of the Generation Floor and Component Importance
Concentration Before Generation-Floor Removal}

Prior to generation floor correction, we observed a substantial generation floor pattern emerge across the full suite of models tested within the cohort, showing a high level of component importance generally centred on the first two and last two transformer blocks of each model. This pattern held regardless of which entity the prompt asked about, with the identity of the dominant component fixed across 68 to 100 \% of prompts in every model. Eleven of the thirteen models place their largest share of effect on the first transformer block, and the two Phi-family models place it on the last. Raw block-lane concentration reaches 0.62 across the cohort (Fig.~\ref{fig:lane-heatmap}). We hypothesise that this reflects where those blocks sit in the residual stream rather than what they contribute to any particular entity. The first block operates on representations that are still close to the raw token embeddings and its output feeds every block that follows, so its removal compounds through the whole network before reaching the answer anchor. The last block makes one of the final writes into the residual stream before the unembedding projection, so its removal arrives at the anchor with nothing downstream to absorb it. In both cases the size of the effect follows from position, and would be expected for any prompt the model is asked to complete.

\subsection{Component Importance Concentration After Generation-Floor Removal}

Once the generation floor is removed, we find no effective evidence that support for a named entity concentrates in a small set of components in any general-purpose model we tested. Removing the floor set strips out the components that carry effect for any short answer, so what remains is the residual attributable to prompt content, and that residual is close to evenly spread. Median floor-removed block-lane LLI sits between 0.017 and 0.035 on medical prompts for eleven of the thirteen models, placing $N_{\mathrm{eff}}/B'$ between 0.46 and 0.82 (Table~\ref{tab:results-combined-no-gemma}, Fig.~\ref{fig:neff-ranked}). We observe two outlier models that do not however follow this trend. First,   MediPhi-Clinical which reaches 0.354 with an effective support of 2.7 components while Phi-4-mini, which shares its architecture and block count, sits at 0.019, so the difference is the clinical fine-tune. Second, Qwen3-1.7B reaches 0.170, and although its low correctness of 35 of 90 is the natural explanation, the table does not support it, since Llama-3.2-1B answered 46 and is the most distributed model in the cohort. What distinguishes Qwen3-1.7B is that its profiles do not change when the entity does, at $-0.007$ with $p = 0.983$. 

\subsection{Component Reuse Across Prompts}

Profile overlap is high everywhere and barely responds to which entity a prompt asks about, so the components carrying effect are largely the same set regardless of content. Because shares are normalised within a lane, two prompts recruiting genuinely different components would produce disjoint mass and a low overlap, and that is not what we observe. Median within-entity overlap runs from 0.512 to 0.906 and across-entity overlap from 0.385 to 0.875, leaving differences of only 0.026 to 0.127 (Table~\ref{tab:results-combined-no-gemma}). Twelve of the thirteen models reach $p < .001$ on that difference, but the effect sizes are small enough that statistical significance is the weaker of the two readings, and pairwise comparisons drawn from a shared prompt set are not independent. Two models sit apart from the rest for opposite reasons. MediPhi-Clinical has the lowest within-entity overlap at 0.512 and the largest separation at 0.127, which is the closest any model comes to profiles that track the entity, while Ministral-3B has the highest overlap in the cohort at 0.906 and 0.865 despite answering none of the 90 prompts correctly, so its near-identical profiles reflect a repeated non-answer rather than shared support. For the question of whether entity knowledge is localised, high overlap that does not move with the entity means that even where effect is concentrated, the concentration is not entity-specific.

%% file: subpages/006_conclusion.tex
We did not observe entity-specific localisation in any instruct-tuned model we tested. The concentration visible before correction belongs to a small set of positionally fixed components that carry effect for any short answer, and once those effects are removed the support that remains is close to evenly spread throughout the model, with profiles that barely change when the named entity does. This holds across every registered lane, from whole blocks down to attention and MLP branches, heads, query groups, positional encodings, attention scores, projections, individual neurons and channels, and output logits. The finding is bounded by what the method measures, namely single-component necessity under one ablation operator read at a single teacher-forced position, and by a general corpus retained on the basis of model correctness. The practical implication is that methods assuming entity knowledge sits in a findable place, including targeted knowledge editing, should first establish that the concentration they observe is not the generation floor.

%% file: subpages/011-Acknowledgment.tex
This research was supported by the Intramural Research Program of the National Institutes of Health (NIH). The contributions of the NIH author(s) are considered Works of the United States Government. The findings and conclusions presented in this paper are those of the author(s) and do not necessarily reflect the views of the NIH or the U.S. Department of Health and Human Services. This work utilized the computational resources of the NIH HPC Biowulf cluster (https://hpc.nih.gov).

%% file: main.bbl
\begin{thebibliography}{11}
\providecommand{\natexlab}[1]{#1}

\bibitem[{Conmy et~al.(2023)Conmy, Mavor-Parker, Lynch, Heimersheim, and Garriga-Alonso}]{3666122.3666841}
Conmy, A.; Mavor-Parker, A.~N.; Lynch, A.; Heimersheim, S.; and Garriga-Alonso, A. 2023.
\newblock Towards automated circuit discovery for mechanistic interpretability.
\newblock In \emph{Proceedings of the 37th International Conference on Neural Information Processing Systems}, NIPS '23. Red Hook, NY, USA: Curran Associates Inc.

\bibitem[{Gallagher et~al.(2024)Gallagher, Ratchford, Brooks, Brown, Heim, McMillan, Nichols, Rallapalli, Smith, VanHoudnos, Winski, and Mellinger}]{gallagher2024assessing}
Gallagher, S.~K.; Ratchford, J.; Brooks, T.; Brown, B.; Heim, E.; McMillan, S.; Nichols, W.~R.; Rallapalli, S.; Smith, C.~J.; VanHoudnos, N.~M.; Winski, N.; and Mellinger, A.~O. 2024.
\newblock Assessing {LLMs} for High Stakes Applications.
\newblock In \emph{Proceedings of the 46th IEEE/ACM International Conference on Software Engineering: Software Engineering in Practice (ICSE-SEIP '24)}, 103--105. Lisbon, Portugal: Association for Computing Machinery.
\newblock ISBN 9798400702174.

\bibitem[{Geva et~al.(2023)Geva, Bastings, Filippova, and Globerson}]{geva2023dissecting}
Geva, M.; Bastings, J.; Filippova, K.; and Globerson, A. 2023.
\newblock Dissecting Recall of Factual Associations in Auto-Regressive Language Models.
\newblock In Bouamor, H.; Pino, J.; and Bali, K., eds., \emph{Proceedings of the 2023 Conference on Empirical Methods in Natural Language Processing}, 12216--12235. Singapore: Association for Computational Linguistics.

\bibitem[{Hirschman(1964)}]{hirschman1964paternity}
Hirschman, A.~O. 1964.
\newblock The Paternity of an Index.
\newblock \emph{The American Economic Review}, 54(5): 761--762.

\bibitem[{Huang et~al.(2025)Huang, Yu, Ma, Zhong, Feng, Wang, Chen, Peng, Feng, Qin, and Liu}]{huang2025hallucination}
Huang, L.; Yu, W.; Ma, W.; Zhong, W.; Feng, Z.; Wang, H.; Chen, Q.; Peng, W.; Feng, X.; Qin, B.; and Liu, T. 2025.
\newblock A Survey on Hallucination in Large Language Models: Principles, Taxonomy, Challenges, and Open Questions.
\newblock \emph{ACM Transactions on Information Systems}, 43(2): 1--55.

\bibitem[{Li et~al.(2024)Li, Li, Song, Yang, Ma, and Yu}]{li2024pmet}
Li, X.; Li, S.; Song, S.; Yang, J.; Ma, J.; and Yu, J. 2024.
\newblock PMET: Precise Model Editing in a Transformer.
\newblock In \emph{Proceedings of the AAAI Conference on Artificial Intelligence}, volume~38, 18564--18572. Association for the Advancement of Artificial Intelligence.

\bibitem[{Meng et~al.(2022)Meng, Bau, Andonian, and Belinkov}]{meng2022locating}
Meng, K.; Bau, D.; Andonian, A.; and Belinkov, Y. 2022.
\newblock Locating and Editing Factual Associations in {GPT}.
\newblock In Koyejo, S.; Mohamed, S.; Agarwal, A.; Belgrave, D.; Cho, K.; and Oh, A., eds., \emph{Advances in Neural Information Processing Systems}, volume~35.
\newblock NeurIPS 2022.

\bibitem[{O'Berry and Kanewala(2026)}]{oberry2026quality}
O'Berry, J.; and Kanewala, U. 2026.
\newblock Quality Assurance of Large Language Model-based Software: A Systematic Literature Review.
\newblock Preprint, Social Science Research Network (SSRN).

\bibitem[{Rai et~al.(2024)Rai, Zhou, Feng, Saparov, and Yao}]{rai2024mechanistic}
Rai, D.; Zhou, Y.; Feng, S.; Saparov, A.; and Yao, Z. 2024.
\newblock A Practical Review of Mechanistic Interpretability for Transformer-Based Language Models.
\newblock \emph{arXiv preprint arXiv:2407.02646}.

\bibitem[{Templeton et~al.(2024)Templeton, Conerly, Marcus, Lindsey, Bricken, Chen, Pearce, Citro, Ameisen, Jones, Cunningham, Turner, McDougall, MacDiarmid, Tamkin, Durmus, Hume, Mosconi, Freeman, Sumers, Rees, Batson, Jermyn, Carter, Olah, and Henighan}]{Templeton2024Scaling}
Templeton, A.; Conerly, T.; Marcus, J.; Lindsey, J.; Bricken, T.; Chen, B.; Pearce, A.; Citro, C.; Ameisen, E.; Jones, A.; Cunningham, H.; Turner, N.~L.; McDougall, C.; MacDiarmid, M.; Tamkin, A.; Durmus, E.; Hume, T.; Mosconi, F.; Freeman, C.~D.; Sumers, T.~R.; Rees, E.; Batson, J.; Jermyn, A.; Carter, S.; Olah, C.; and Henighan, T. 2024.
\newblock Scaling Monosemanticity: Extracting Interpretable Features from Claude 3 Sonnet.
\newblock \url{https://transformer-circuits.pub/2024/scaling-monosemanticity/}.
\newblock Transformer Circuits Thread.

\bibitem[{Zhao et~al.(2024)Zhao, Chen, Yang, Liu, Deng, Cai, Wang, Yin, and Du}]{zhao2024explainability}
Zhao, H.; Chen, H.; Yang, F.; Liu, N.; Deng, H.; Cai, H.; Wang, S.; Yin, D.; and Du, M. 2024.
\newblock Explainability for Large Language Models: A Survey.
\newblock \emph{ACM Transactions on Intelligent Systems and Technology}, 15(2): 1--38.

\end{thebibliography}
